%% file: paper.tex
\documentclass{article}
\usepackage{iclr2027_conference,times}
\iclrfinalcopy

\makeatletter
\@ifundefined{iclrfinalcopy}{%
    \PackageError{MorphAct}{ICLR style is empty or invalid}{%
        Restore the complete iclr2027_conference.sty file.%
    }\@@end
}{}
\@ifundefined{citealp}{%
    \PackageError{MorphAct}{natbib did not load}{%
        Restore the complete template dependencies.%
    }\@@end
}{}
\makeatother

\usepackage[T1]{fontenc}
\usepackage{amsmath,amssymb}
\usepackage{graphicx,booktabs,tabularx,array,multirow,colortbl,siunitx}
\usepackage{svg}
\usepackage{xcolor}
\usepackage{microtype}
\usepackage{needspace}
\usepackage{fontawesome5}

\usepackage{algorithm}
\usepackage{algpseudocode}

\algrenewcommand{\algorithmicrequire}{\textbf{Input:}}
\algrenewcommand{\algorithmicensure}{\textbf{Output:}}

\usepackage{hyperref}

\hypersetup{
    colorlinks=false,
    pdfborder={0 0 1},
    citebordercolor={0 1 0},
    linkbordercolor={1 0 0},
    urlbordercolor={0 0 1},
    pdfauthor={Ziyan Liu, Jiting Liu, Weihong Zhong, Wenda Yu, Yeqiu Chen, Jason Li, Xiaojiang Peng, Shuyuan Hu, Kai Wang, Bo Zhao}
}

\definecolor{PendingColor}{RGB}{130,75,15}
\definecolor{MorphActRow}{RGB}{239,245,239}
\definecolor{TitleLinkBlue}{RGB}{31,90,166}
\renewcommand{\theequation}{\arabic{equation}}

\DeclareRobustCommand{\morphicon}{%
    \raisebox{-0.25em}[0.9em][0pt]{%
        \includesvg[
            height=2.2em,
            width=1.9em,
            keepaspectratio=true,
            inkscapearea=drawing,
            inkscapelatex=false
        ]{figures/morphact-icon-v2.svg}%
    }%
}

\hypersetup{
    colorlinks=true,
    linkcolor=red,
    citecolor=blue,
    filecolor=magenta,      
    urlcolor=magenta,
    }
\title{%
    \texorpdfstring{\morphicon\hspace{0.20em}}{}%
    MorphAct: Robot Policies Morphing with the Worlds They Act In%
}

\author{%
    \hspace*{-\tabcolsep}%
    \parbox[t]{\dimexpr\textwidth-\tabcolsep\relax}{%
        \raggedright\normalfont
        \textbf{%
            Ziyan Liu\textsuperscript{*\,1,2}\quad
            Jiting Liu\textsuperscript{*\,2}\quad
            Weihong Zhong\textsuperscript{*\,2}\quad
            Wenda Yu\textsuperscript{5}\quad
            Yeqiu Chen\textsuperscript{1}}\\[2pt]
        \textbf{%
            Jason Li\textsuperscript{4}\quad
            Xiaojiang Peng\textsuperscript{6}\quad
            Shuyuan Hu\textsuperscript{2}\quad
            Kai Wang\textsuperscript{3\,\textdagger}\quad
            Bo Zhao\textsuperscript{2\,\textdagger}}\\[5pt]
        {\normalsize
            \begin{tabularx}{\linewidth}{@{}>{\raggedright\arraybackslash}X@{\quad}>{\raggedright\arraybackslash}X@{}}
                \multicolumn{2}{@{}l@{}}{\textsuperscript{1}University of Science and Technology of China}\\[1pt]
                \multicolumn{2}{@{}l@{}}{\textsuperscript{2}School of Artificial Intelligence, Shanghai Jiao Tong University}\\[1pt]
                \textsuperscript{3}National University of Singapore &
                \textsuperscript{4}Nanyang Technological University\\[1pt]
                \textsuperscript{5}Tongji University &
                \textsuperscript{6}Shenzhen Technology University
            \end{tabularx}\\[3pt]}
        {\normalsize\hypersetup{hidelinks}
            \textsuperscript{*}Equal contribution\qquad
            \textsuperscript{\textdagger}Corresponding authors\qquad
            \href{https://mint-sjtu.github.io/MorphAct.io/}{\textcolor{TitleLinkBlue}{\faGlobe\enspace Project page}}\endgraf}%
    }%
}

\begin{document}

\maketitle
% Keep the camera-ready header placement without claiming conference acceptance.
\lhead{MorphAct}

\input{sections/abstract}
\input{sections/teaser}
\input{sections/introduction}
\input{sections/motivation}
\input{sections/method}
\input{sections/experiments}
\input{sections/analysis}
\input{sections/related_work}
\input{sections/conclusion}
\input{sections/acknowledgments}

\bibliographystyle{iclr2027_conference}
\bibliography{ref_clean}

\input{sections/appendix}

\end{document}

%% file: sections/abstract.tex
% Section source. See project README for the local-to-Overleaf editing workflow.
\begin{abstract}

Progress toward general intelligence in the physical world requires scaling robot learning across diverse environments and interactions. However, existing robot policies use shared parameters that remain fixed across interaction contexts. When different operations favor competing parameter adjustments, fitting their mixture can leave individual operations insufficiently adapted. Our atomic-task diagnostics reveal competing update directions in both vision--language and action parameters, with source-improving updates increasing loss on other tasks. This motivates \textbf{MorphAct}\footnote{\normalsize ``Morph,'' rooted in Greek \emph{morph\={e}} (``form''), means ``to transform.''}, a contextual parameterization framework that dynamically adjusts robot policy parameters as interaction contexts change. Two coupled generators produce context-dependent low-rank updates for the vision--language and action modules. Their coupling follows the policy's computational dependency: vision--language updates shape the representations that condition action-side parameter generation. Both generators are trained jointly through the native action objective, without operation labels or expert-adapter targets.
MorphAct improves performance across three VLA backbones on LIBERO and Meta-World, with gains of up to 16.30\%. In real-world evaluation, it achieves gains of up to 14\% over $\pi_{0.5}$. Ablations support the benefits of contextual parameter generation and representation coupling.

\end{abstract}

%% file: sections/teaser.tex
% Section source. See project README for the local-to-Overleaf editing workflow.
\suppressfloats[t]
\begin{figure}[t]
\centering
\includegraphics[width=\linewidth]{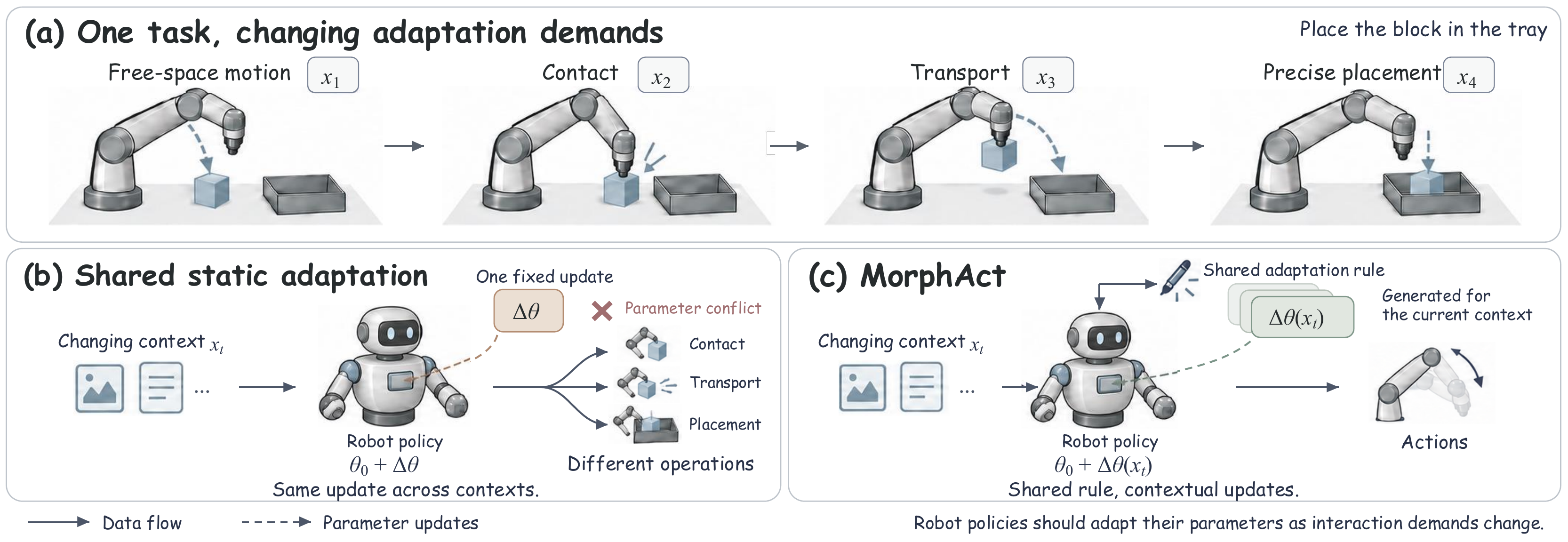}
\caption{
\textbf{Robot policies morph with changing interaction contexts.}
(a) A single manipulation task involves operations with different
perceptual and control demands.
(b) Existing robot policies use one fixed parameter configuration across
contexts, requiring shared parameters to reconcile competing demands.
(c) MorphAct generates context-dependent updates for the vision--language
and action modules, dynamically adjusting policy parameters as interactions unfold.
}
\label{fig:morphact-teaser}
\end{figure}

%% file: sections/introduction.tex
% Section source. See project README for the local-to-Overleaf editing workflow.
\section{Introduction}

Progress toward general intelligence in the physical world requires scaling robot learning across diverse environments, operations, and interactions \citep{driess2023palme,black2025pi0}. This expansion brings together heterogeneous perceptual and control demands within a shared model. Such diversity arises not only across tasks but also within a single interaction: placing an object in a tray involves approaching it, establishing contact, transporting it, and positioning it precisely. These operations require different visual cues and control behaviors, often under an unchanged instruction. As environmental and operational coverage expands, a central challenge is how a shared model can effectively accommodate these heterogeneous demands.

Vision--language--action (VLA) models provide a promising foundation by combining pretrained multimodal representations with action prediction \citep{brohan2023rt2,kim2025openvla,black2025pi0,liu2025bridging}. However, conventional VLA policies execute diverse interactions using a shared, static parameter configuration. Context changes their activations and predicted actions, but their effective weights remain fixed, including after full fine-tuning or low-rank adaptation \citep{hu2022lora,kim2025oft}. Different operations must therefore be learned through shared parameters, even when their training signals favor competing updates---a source of interference in joint optimization \citep{yu2020pcgrad}. This raises the question of how effectively shared parameters reconcile the demands of different operations.

We investigate this question through operation-level diagnostics using demonstrations from the 65 atomic tasks in RoboCasa365 \citep{nasiriany2026robocasa365}. At a common policy checkpoint, we examine gradient alignment, the effect of a single update on other tasks, and the contributions of individual parameter groups. The diagnostics reveal competing update directions in both the vision--language and action modules. Every measured source-improving update also increases loss on batches from other tasks, while opposing parameter-group contributions can cancel within a module. These findings expose local adaptation trade-offs in shared parameter space. They motivate a shared parameter-generation mechanism whose outputs vary with the interaction context.

To address this challenge, we introduce \textbf{MorphAct}, a contextual parameterization framework that enables robot policies to \emph{morph with changing interaction contexts} (Figure~\ref{fig:morphact-teaser}). MorphAct generates context-dependent low-rank parameter updates for the vision--language and action modules, allowing a shared policy to adjust its effective weights to the current interaction. Because vision--language adaptation changes the representations used for control, MorphAct couples the two generators by conditioning action-side parameter generation on the adapted representations. Both generators learn jointly through the native action objective, without operation labels or expert-adapter targets. Low-rank factors provide an efficient implementation of these contextual updates.

We evaluate MorphAct on X-VLA, $\pi_0$, and $\pi_{0.5}$ \citep{zheng2026xvla,black2025pi0,intelligence2025pi05} using LIBERO \citep{liu2023libero} and Meta-World \citep{yu2020metaworld}, comparing each backbone's baseline with MorphAct. Across the three backbones, MorphAct improves Meta-World average success by 7.57--16.30\%; on LIBERO, X-VLA, $\pi_0$, and $\pi_{0.5}$ gain 4.15\%, 2.50\%, and 1.75\%, respectively. Real-world evaluation with $\pi_{0.5}$ further yields gains of 8.00--14.00\% across four manipulation tasks. Controlled X-VLA ablations on LIBERO show that fixed generation conditions, single-module adaptation, and action-side generation conditioned on pre-adaptation features each reduce mean success. These results support contextual parameterization and representation coupling for accommodating diverse manipulation demands.

Our contributions are threefold: \textbf{(1)}~We identify parameter conflicts as a challenge to scaling robot learning across diverse environments and interactions through operation-level diagnostic experiments. \textbf{(2)}~We introduce MorphAct, whose representation-coupled generators produce context-dependent vision--language and action parameter updates and learn jointly through the native action objective. \textbf{(3)}~We demonstrate success gains on three VLA backbones, two simulation benchmarks, and four real-world tasks, with ablations supporting contextual generation and representation coupling.

%% file: sections/motivation.tex
% Section source. See project README for the local-to-Overleaf editing workflow.
\section{Parameter Conflicts Across Operations}
\label{sec:diagnosis}

As robot learning expands across environments and interactions,
static-parameter policies must reconcile diverse operational demands
through the same weights. We examine whether different operations
favor compatible changes to these parameters. Our diagnostics measure
gradient alignment, one-step cross-task loss changes, and parameter-group
contributions in both vision--language and action modules at a common
checkpoint.

\textbf{Setup.}
We use demonstrations from the 65 atomic tasks in RoboCasa365
\citep{nasiriany2026robocasa365}, grouped by native task ID.
All measurements use a fixed RoboCasa-pretrained $\pi_{0.5}$ checkpoint
and the native action loss $\mathcal L$ over the full vision--language
(VLM) and action parameter sets.
Each sampling round draws batches $U_i$ and $V_i$ from disjoint
episodes of task $i$.
Gradient comparisons use $(U_i,U_j)$ across tasks and $(U_i,V_i)$
as a within-task reference.
For one-step transfer, $U_i$ supplies the update and $V_j$ the target evaluation.
Appendix~\ref{app:diagnostics} details sampling, numerical settings, and metrics.

\begin{figure}[!t]
    \centering
    \includegraphics[width=\linewidth]{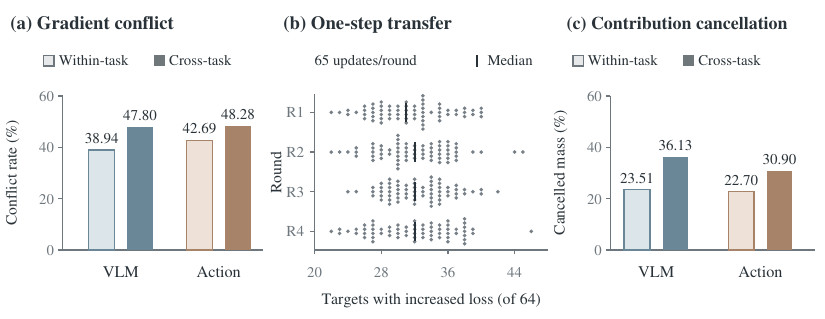}
    \caption{
    \textbf{Parameter conflicts across operations and within modules.}
    (a) Negative-gradient-cosine frequency within and across tasks.
    (b) Number of other tasks with increased target-batch loss after a source update.
    All 260 updates reduce source-batch loss; dots represent updates
    and ticks mark round medians.
    (c) Fraction of absolute alignment contribution cancelled across
    parameter groups within each module.
    Measurements use one checkpoint with 16 sampling rounds for (a,c)
    and four for (b).
    }
    \label{fig:adaptation-diagnostics}
\end{figure}

\subsection{Conflicting Directions in Both Modules}
\label{sec:module-interference}

We first ask whether different tasks favor compatible parameter changes.
At the common checkpoint, we compute task-batch gradients separately
for the VLM and action modules and compare their cosine similarity.
A negative cosine indicates competing local descent directions
\citep{yu2020pcgrad}.
Across 16 sampling rounds, negative alignment occurs in 47.80\% of
cross-task VLM comparisons and 48.28\% of action comparisons,
versus 38.94\% and 42.69\% within tasks
(Figure~\ref{fig:adaptation-diagnostics}(a)).
The within-task reference shows that update directions also vary
among episodes of the same task, while cross-task comparisons
produce negative alignment more frequently.
Competing adaptation demands therefore affect both the
representations used for control and the action mapping.

\subsection{Cross-Task Interference from Parameter Updates}
\label{sec:one-step-transfer}

Gradient alignment describes local update directions; we next measure
the loss changes produced by an actual update.
For each source batch $U_i$, we take one full-parameter gradient step
and evaluate the target batches $V_j$, restarting from the common
checkpoint for every source.
Target observations, action targets, flow times, and noise remain
fixed before and after the step.
Across four rounds, all 260 updates reduce their source-batch loss,
with a mean relative reduction of 2.38\%.
Yet each update increases loss on batches from 22--46 of the other
64 tasks, with a median of 32
(Figure~\ref{fig:adaptation-diagnostics}(b)).
Every source-improving update thus has both helpful and harmful
effects on other tasks: improving the source does not produce
uniformly beneficial transfer across manipulation contexts.

\subsection{Conflicting Contributions Across Parameter Groups}
\label{sec:contribution-cancellation}

The preceding comparisons summarize each module as a whole.
To examine the interactions within these summaries, we decompose
gradient alignment across 47 VLM groups and 19 action groups,
including individual Transformer blocks \citep{shi2023recon}.
Each group's gradient inner product is normalized by the same
full-module gradient norms, so the contributions sum to the
module-level cosine.
Positive and negative contributions can offset one another.
We sum the absolute contribution mass canceled within each pair
and divide by the total absolute contribution mass across comparisons.
For cross-task pairs, cancellation reaches 36.13\% in the VLM
and 30.90\% in the action module, compared with 23.51\% and
22.70\% within tasks
(Figure~\ref{fig:adaptation-diagnostics}(c)).
A small module-level cosine can therefore reflect opposing
parameter-group contributions rather than uniformly weak interactions.
Appendix~\ref{app:diagnostic-groups} reports the complete group
definitions and decomposition.

\textbf{Implication for MorphAct.}
Together, these results show that adaptation demands differ across
tasks and act on both representation and action parameters.
An update that serves one context can work against another,
with opposing contributions hidden by module-level summaries.
This motivates sharing the adaptation mechanism while allowing
its updates to depend on the current context.
MorphAct uses two generators connected
by the adapted visual--language representation.

%% file: sections/method.tex
% Section source. See project README for the local-to-Overleaf editing workflow.
% Replace numerical equation references elsewhere with the labels defined below.
\section{MorphAct: Morph with the Environment and Then Act}
\label{sec:morphact}

\begin{figure}[!t]
\centering
\includegraphics[width=\linewidth]{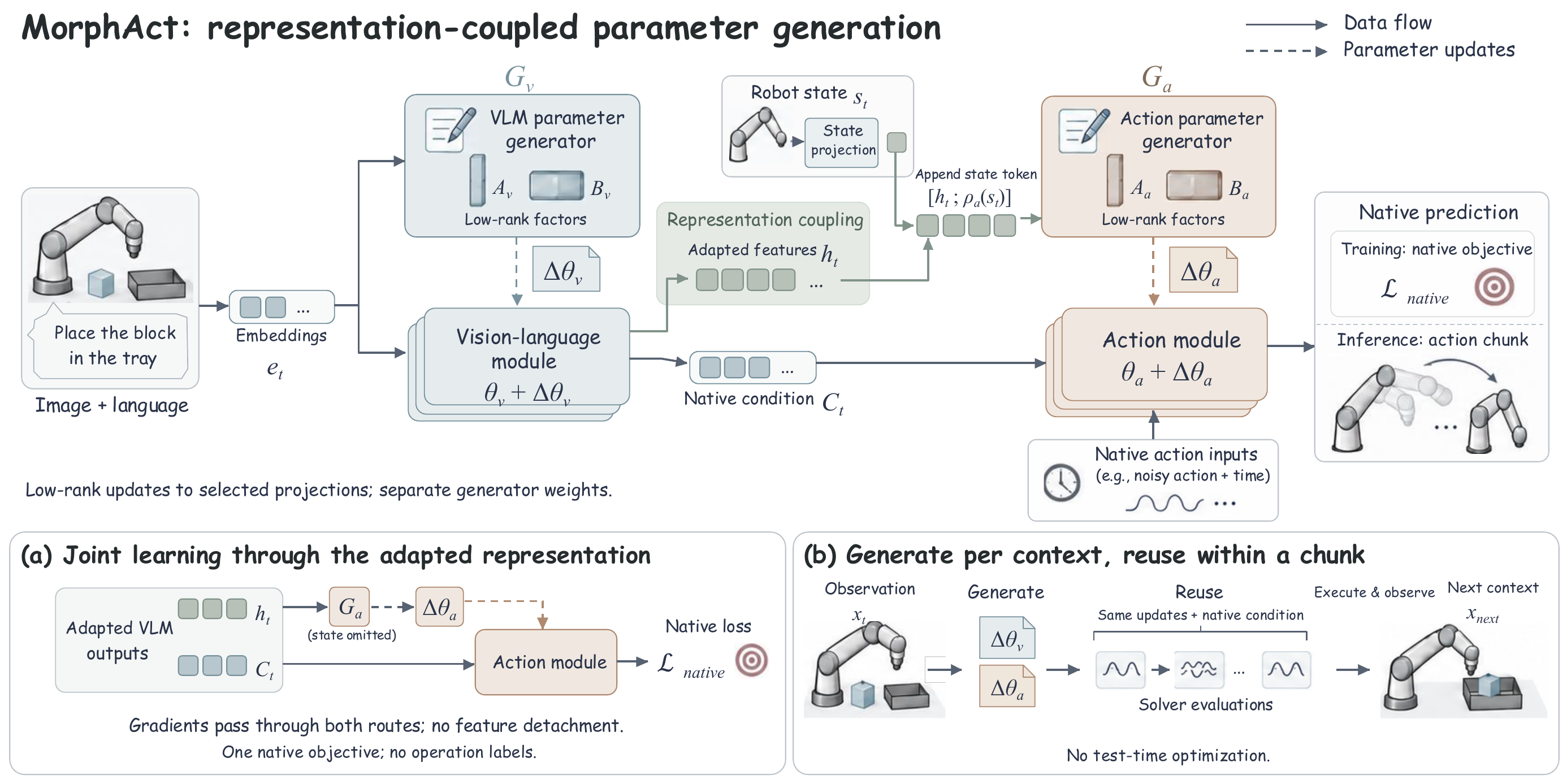}
\caption{
\textbf{Representation-coupled parameter generation.}
$G_v$ generates vision--language updates from multimodal embeddings.
The adapted VLM pass produces $h_t$ for action-side parameter generation
and $C_t$ for native action conditioning.
$G_a$ generates action updates from $h_t$ and robot state.
Solid arrows carry data; dashed arrows inject updates.
Both generators learn through the action objective, and their outputs
are reused within each action chunk.
}
\label{fig:morphact-overview}
\end{figure}

MorphAct adjusts robot policy parameters to the current interaction context before predicting an action chunk (Figure~\ref{fig:morphact-overview}). Two generators produce context-dependent updates for the vision--language and action modules, with action-side generation conditioned on the adapted vision--language representations. We describe this coupled generation process and its low-rank parameterization, then explain how both generators learn through the native action objective and how their outputs are reused within each action chunk.

\subsection{Representation-Coupled Generation}
\label{sec:morphact-generation}

Let $x_t=(o_t,\ell,s_t)$ denote the observation, instruction, and robot state at replanning time $t$. We denote the base policy parameters by $\theta$, with vision--language and action modules parameterized by $\theta_v$ and $\theta_a$. MorphAct generates $\Delta\theta(x_t)=(\Delta\theta_{v,t},\Delta\theta_{a,t})$ for selected projections and predicts an action chunk as $\widehat Y_t=\pi_{\theta+\Delta\theta(x_t)}(x_t)$, suppressing sampling randomness. The generator weights are shared across contexts, while their outputs determine the context-dependent effective parameters.

The two updates follow the policy's computational dependency: vision--language adaptation changes the representations used for control, so action-side generation conditions on the adapted features. Given multimodal embeddings $e_t=E(o_t,\ell)$, the forward computation is
\begin{equation}
\begin{alignedat}{2}
\Delta\theta_{v,t}
&=G_v(e_t;\phi_v),\qquad &
(h_t,C_t)
&=F_v(e_t;\theta_v+\Delta\theta_{v,t}),\\
u_{a,t}
&=C_a(h_t,s_t;\rho_a), &
\Delta\theta_{a,t}
&=G_a(u_{a,t};\phi_a).
\end{alignedat}
\label{eq:morphact-generation}
\end{equation}
Here, $h_t$ supplies the action generator's visual--language condition, whereas $C_t$ retains the action module's native conditioning path. Both are outputs of the same adapted VLM pass. In $\pi_0$ and $\pi_{0.5}$, they correspond to final multimodal prefix tokens and layer-wise prefix key--value states, respectively. The constructor $C_a$ appends a robot-state token projected by $\rho_a$ to $h_t$. Thus, the generators have separate parameters but are connected through the representation produced by upstream adaptation.

\textbf{Structured low-rank synthesis.}
Both generators use the structured synthesis primitive of HY-WU \citep{wu2026hy}, with distinct target layouts for the two modules. Queries encode layer and projection/rank identities; generator blocks combine cross-layer attention, within-layer parameter attention, and cross-attention to the condition tokens. Separate A/B readouts then produce the factors for each selected block. The generators attend to individual condition tokens.

For block $j$, the generated factors are $A_{j,t}\in\mathbb R^{d_{\mathrm{in},j}\times r_j}$ and $B_{j,t}\in\mathbb R^{r_j\times d_{\mathrm{out},j}}$. With scaling coefficient $\alpha_j$, the update and its residual contribution for a row-vector input $u$ are
\begin{equation}
\Delta W_{j,t}
=
\frac{\alpha_j}{r_j}A_{j,t}B_{j,t},
\qquad
\delta y_{j,t}
=
\frac{\alpha_j}{r_j}(uA_{j,t})B_{j,t}.
\label{eq:morphact-low-rank}
\end{equation}
The factored computation avoids materializing dense updates. Q/K/V projections use output-head blocks with their actual query and key/value head counts; attention-output projections sum input-head contributions, while feed-forward projections use matrix-level factors. These layouts accommodate the modules' different widths and depths. The rank $r_j$ is defined per generated block, not for the assembled multi-head projection. Appendix~\ref{app:parameter-geometry} specifies the factor geometry and output budgets.

\subsection{Learning through the Adapted Representation}
\label{sec:morphact-learning}

MorphAct trains both generators jointly through the base policy's action objective, using the same adaptation targets as the corresponding LoRA configuration. Let $\Phi=(\phi_v,\phi_a,\rho_a,\psi)$ denote the generator parameters, state projection, and projector parameters $\psi$ retained as trainable in that configuration. The remaining VLM and action backbone parameters are frozen. Generated factors are differentiable intermediate outputs, not independent optimization variables. For demonstrations $(x_t,Y_t)\sim\mathcal D$, we optimize
\begin{equation}
\min_{\Phi}\;
\mathbb E_{(x_t,Y_t)\sim\mathcal D,\,\xi}
\left[
\mathcal L_{\mathrm{native}}
\bigl(\widehat y_\Phi(x_t,Y_t;\xi),Y_t;\xi\bigr)
\right],
\label{eq:morphact-objective}
\end{equation}
where $\xi$ contains the randomness required by the native training objective, and $\widehat y_\Phi$ denotes its training-time prediction. For $\pi_0$ and $\pi_{0.5}$, the sampled time and noise $\xi=(\tau,\epsilon)$ define $z_\tau=\tau\epsilon+(1-\tau)Y_t$, and the model predicts the velocity $\epsilon-Y_t$, rather than the clean inference chunk $\widehat Y_t$. This is conditional flow matching along a linear action--noise path \citep{lipman2023flowmatching}.

The adapted representation participates in learning through two routes: native action conditioning and action-side parameter generation. Writing $r_t=(h_t,C_t)$ and the per-sample loss as $\ell_t(r_t,\Delta\theta_{a,t})$, with other inputs suppressed, the upstream gradient is
\begin{equation}
\nabla_{\phi_v}\ell_t
=
J_{r_t,\phi_v}^{\top}
\Bigl[
\frac{\partial\ell_t}{\partial r_t}
+
J_{\Delta\theta_{a,t},r_t}^{\top}
\frac{\partial\ell_t}{\partial\Delta\theta_{a,t}}
\Bigr],
\qquad
J_{y,z}=\frac{\partial y}{\partial z}.
\label{eq:morphact-coupling-gradient}
\end{equation}
The first term differentiates through action conditioning while holding the action update fixed; the second accounts for the update's dependence on adapted features. Neither route is detached. The action objective therefore trains the upstream generator through both the representations used for control and the parameters generated from those representations.

\subsection{Context-Level Adaptation, Chunk-Level Reuse}
\label{sec:morphact-execution}

Action-sequence prediction underlies both ACT's action chunking and Diffusion Policy's receding-horizon control \citep{zhao2023act,chi2023diffusionpolicy}.
Parameter generation depends on the current observation context, not on noisy actions or internal solver time. At each replanning step, MorphAct generates the VLM update, computes $h_t$ and $C_t$, and generates the action update. These quantities are then reused throughout action-chunk prediction and refreshed at the next replanning observation, without test-time optimization. In $\pi_0$ and $\pi_{0.5}$, the single adapted prefix pass supplies both representation outputs, while the action expert retains its native state and time interfaces, including $\pi_{0.5}$'s time-conditioned normalization. Appendix~C provides the inference procedure.

\subsection{Implementation Details}
\label{sec:morphact-implementation}

Both A/B readouts use small, nonzero weight initialization and zero
biases, keeping initial residuals small without blocking gradients
to the generator trunk.
Generator parameters are stored in FP32, while internal attention
and feed-forward layers support BF16 computation.
Factor readouts, low-rank products, scaling, and residual accumulation
are computed in FP32 before conversion to the backbone output dtype.
Training uses AdamW with learning-rate warmup, cosine decay, and
global gradient clipping; updates with non-finite losses or gradient
norms are skipped.

For variable-length conditions, valid tokens are compacted using
the supplied masks and grouped by sequence length.
The original batch order is restored after generation, preserving
gradients through the adapted features.
Non-reentrant activation checkpointing reduces training memory.
Algorithms~\ref{alg:morphact-training} and
\ref{alg:morphact-inference} in
Appendix~\ref{app:training-inference} provide the complete training
and inference procedures, respectively.

%% file: sections/experiments.tex
% Section source. See project README for the local-to-Overleaf editing workflow.
\section{Experiments}
\label{sec:experiments}

We evaluate MorphAct on LIBERO \citep{liu2023libero},
Meta-World \citep{yu2020metaworld}, and four real-world manipulation
tasks.
These closed-loop evaluations complement the atomic-task analysis
in Section~\ref{sec:diagnosis} by measuring complete task execution.
Section~\ref{sec:analysis} examines the contributions of individual
design choices through ablations.

\subsection{Evaluation Setting}
\label{sec:eval-setting}
We use closed-loop task success as the primary metric. In simulation,
we compare MorphAct with the corresponding baselines for X-VLA,
$\pi_0$, and $\pi_{0.5}$ under the same task settings.
On LIBERO, we train jointly on Spatial, Object, Goal, and Long
(LIBERO-10), each containing ten tasks, and report suite-level success
and their arithmetic average as the Mean score.
On Meta-World, we train jointly on all 50 tasks in MT50 and report
success for four difficulty groups: Easy (28 tasks), Medium (11),
Hard (6), and Very Hard (5). Avg. is the unweighted arithmetic mean
of the four group success rates.

We evaluate 50 rollouts per task on LIBERO and twenty on Meta-World.
MorphAct uses execution horizon H10 on LIBERO. On Meta-World,
its checkpoint and execution-horizon settings are 20k / H3 for $\pi_0$,
17.5k / H4 for $\pi_{0.5}$, and 82k / H4 for X-VLA.
The two main simulation tables report MorphAct's mean scores across
three random seeds, each using these rollout counts, and the sample
variance of the three overall scores in squared percentage points
(pp$^2$). Baseline scores are retained from the original comparison;
their variances are not reported.

For real-world evaluation, we compare $\pi_{0.5}$ with and without
MorphAct on four manipulation tasks, using 50 physical rollouts per
task for each method. Section~\ref{sec:real-world} describes the tasks.
\subsection{LIBERO}
\label{sec:manipulation-performance}

Table~\ref{tab:performance} shows that MorphAct improves mean success
on all three backbones. On X-VLA, success rises from 93.05\% to
97.20\%, a gain of 4.15 percentage points. On $\pi_0$, it increases
from 94.15\% to 96.65\%, a gain of 2.50 points. MorphAct also improves
$\pi_{0.5}$ from 96.85\% to 98.60\%, adding 1.75 points to the
highest baseline mean in this comparison.

Long shows the largest gain for each backbone: 9.00 points for X-VLA,
7.20 for $\pi_0$, and 5.13 for $\pi_{0.5}$.
MorphAct improves all four suites on X-VLA and $\pi_0$.
For X-VLA, Goal rises from 96.00\% to 97.07\%.
For $\pi_{0.5}$, it improves Object, Goal, and Long,
while Spatial changes from 98.80\% to 98.33\%.
Thus, the higher mean scores combine a consistent improvement on
Long with more varied changes on the remaining suites.

\begin{table}[htbp]
\centering
\normalsize
\setlength{\tabcolsep}{3pt}
\renewcommand{\arraystretch}{1.15}
\begin{tabular}{
    @{}>{\raggedright\arraybackslash}p{.105\linewidth}
    >{\raggedright\arraybackslash}p{.125\linewidth}
    *{5}{S[
        table-format=2.2,
        detect-weight=true,
        table-column-width=\dimexpr.128\linewidth-2.4\tabcolsep\relax
    ]}
    S[table-format=1.4, table-column-width=\dimexpr.13\linewidth-2\tabcolsep\relax]@{}
}
\toprule
Backbone & Method
& \multicolumn{1}{c}{Spatial $\uparrow$}
& \multicolumn{1}{c}{Object $\uparrow$}
& \multicolumn{1}{c}{Goal $\uparrow$}
& \multicolumn{1}{c}{Long $\uparrow$}
& \multicolumn{1}{c}{Mean $\uparrow$}
& \multicolumn{1}{c}{Var. (pp$^2$)} \\
\midrule
& Baseline
& 95.40 & 96.60 & 96.00 & 84.20 & 93.05 & {--} \\
\rowcolor{MorphActRow}
\cellcolor{white}\multirow{-2}{*}{\textit{X-VLA}} & MorphAct
& \textbf{98.67} & \textbf{99.87} & \textbf{97.07} & \textbf{93.20} & \textbf{97.20} & 0.0300 \\
\midrule
& Baseline
& 96.80 & 98.80 & 95.80 & 85.20 & 94.15 & {--} \\
\rowcolor{MorphActRow}
\cellcolor{white}\multirow{-2}{*}{\textit{$\pi_0$}} & MorphAct
& \textbf{98.00} & \textbf{99.60} & \textbf{96.60} & \textbf{92.40} & \textbf{96.65} & 0.0517 \\
\midrule
& Baseline
& \textbf{98.80} & 98.20 & 98.00 & 92.40 & 96.85 & {--} \\
\rowcolor{MorphActRow}
\cellcolor{white}\multirow{-2}{*}{\textit{$\pi_{0.5}$}} & MorphAct
& 98.33 & \textbf{99.67} & \textbf{98.87} & \textbf{97.53} & \textbf{98.60} & 0.0525 \\
\bottomrule
\end{tabular}
\caption{
\textbf{LIBERO closed-loop success rates (\%).}
Long denotes LIBERO-10; Mean averages the four suite scores.
MorphAct scores are averaged across three random seeds, each covering
40 tasks with 50 rollouts per task and execution horizon H10.
Var. is the sample variance of the three overall scores (pp$^2$);
-- denotes unreported baseline variance.
Shaded rows denote MorphAct. Bold indicates the best result
within each backbone and column, including ties.
}
\label{tab:performance}
\end{table}

\subsection{Meta-World}
\label{sec:metaworld}

Table~\ref{tab:metaworld} reports performance under joint training
on MT50. MorphAct improves average success from 55.54\% to 63.11\%
on X-VLA, from 50.75\% to 67.05\% on $\pi_0$, and from 48.70\%
to 64.42\% on $\pi_{0.5}$.
The corresponding gains are 7.57, 16.30, and 15.72 percentage
points. All three backbones therefore benefit from MorphAct
on this multi-task benchmark.

The improvements span the difficulty groups rather than appearing
only in the aggregate score. $\pi_0$ and $\pi_{0.5}$ improve in all
four groups, while X-VLA improves in three and matches its baseline
on Very Hard at 55.00\%.
Some of the largest gains occur on the harder groups:
$\pi_0$ rises from 20.00\% to 58.00\% on Very Hard, and $\pi_{0.5}$
rises from 38.30\% to 66.39\% on Hard.
For X-VLA, the largest gain is on Medium, from 46.36\%
to 58.18\%. These results show benefits across difficulty levels,
with the distribution of gains depending on the backbone.

\begin{table}[htbp]
\centering
\normalsize
\setlength{\tabcolsep}{3pt}
\renewcommand{\arraystretch}{1.15}
\begin{tabular}{
    @{}>{\raggedright\arraybackslash}p{.105\linewidth}
    >{\raggedright\arraybackslash}p{.125\linewidth}
    S[table-format=2.2, detect-weight=true, table-column-width=\dimexpr.115\linewidth-2.4\tabcolsep\relax]
    S[table-format=2.2, detect-weight=true, table-column-width=\dimexpr.135\linewidth-2.4\tabcolsep\relax]
    S[table-format=2.2, detect-weight=true, table-column-width=\dimexpr.115\linewidth-2.4\tabcolsep\relax]
    S[table-format=2.2, detect-weight=true, table-column-width=\dimexpr.16\linewidth-2.4\tabcolsep\relax]
    S[table-format=2.2, detect-weight=true, table-column-width=\dimexpr.115\linewidth-2.4\tabcolsep\relax]
    S[table-format=1.4, table-column-width=\dimexpr.13\linewidth-2\tabcolsep\relax]@{}
}
\toprule
Backbone & Method
& \multicolumn{1}{c}{Easy $\uparrow$}
& \multicolumn{1}{c}{Medium $\uparrow$}
& \multicolumn{1}{c}{Hard $\uparrow$}
& \multicolumn{1}{c}{Very Hard $\uparrow$}
& \multicolumn{1}{c}{Avg. $\uparrow$}
& \multicolumn{1}{c}{Var. (pp$^2$)} \\
\midrule
& Baseline
& 72.50 & 46.36 & 48.30 & \textbf{55.00} & 55.54 & {--} \\
\rowcolor{MorphActRow}
\cellcolor{white}\multirow{-2}{*}{\textit{X-VLA}} & MorphAct
& \textbf{83.99} & \textbf{58.18} & \textbf{55.28} & \textbf{55.00} & \textbf{63.11} & 1.8400 \\
\midrule
& Baseline
& 77.90 & 51.80 & 53.30 & 20.00 & 50.75 & {--} \\
\rowcolor{MorphActRow}
\cellcolor{white}\multirow{-2}{*}{\textit{$\pi_0$}} & MorphAct
& \textbf{82.44} & \textbf{71.36} & \textbf{56.39} & \textbf{58.00} & \textbf{67.05} & 0.6342 \\
\midrule
& Baseline
& 70.50 & 45.00 & 38.30 & 41.00 & 48.70 & {--} \\
\rowcolor{MorphActRow}
\cellcolor{white}\multirow{-2}{*}{\textit{$\pi_{0.5}$}} & MorphAct
& \textbf{85.77} & \textbf{59.85} & \textbf{66.39} & \textbf{45.67} & \textbf{64.42} & 0.3970 \\
\bottomrule
\end{tabular}
\caption{
\textbf{Meta-World closed-loop success rates (\%).}
Avg. is the unweighted arithmetic mean of the Easy, Medium, Hard,
and Very Hard success rates. MorphAct scores are averaged across three
random seeds, each covering 50 tasks with 20 rollouts per task.
Var. is the sample variance of the three overall scores (pp$^2$);
-- denotes unreported baseline variance.
Shaded rows denote MorphAct. Bold indicates the best result
within each backbone and column, including ties.
}
\label{tab:metaworld}
\end{table}

\begin{figure}[!htbp]
\centering
\newsavebox{\realWorldTableBox}
\newsavebox{\realWorldImagesBox}
\sbox{\realWorldTableBox}{%
\begin{minipage}{.62\linewidth}
\centering
\normalsize
\setlength{\tabcolsep}{2.5pt}
\renewcommand{\arraystretch}{1.35}
\begin{tabularx}{\linewidth}{@{}l*{3}{>{\centering\arraybackslash}X}@{}}
\toprule
Task & Baseline & MorphAct & Gain \\
\midrule
Sort Letters & 82.00\% & \textbf{92.00\%} & +10.00 \\
Sort Hardware & 8.00\% & \textbf{22.00\%} & +14.00 \\
Deliver Water & 78.00\% & \textbf{86.00\%} & +8.00 \\
Cabinet Storage & 84.00\% & \textbf{96.00\%} & +12.00 \\
\bottomrule
\end{tabularx}
\end{minipage}%
}
\sbox{\realWorldImagesBox}{%
\setlength{\tabcolsep}{0pt}%
\renewcommand{\arraystretch}{0}%
\begin{tabular}{@{}c@{\hspace{3pt}}c@{}}
\includegraphics[width=.23\linewidth]{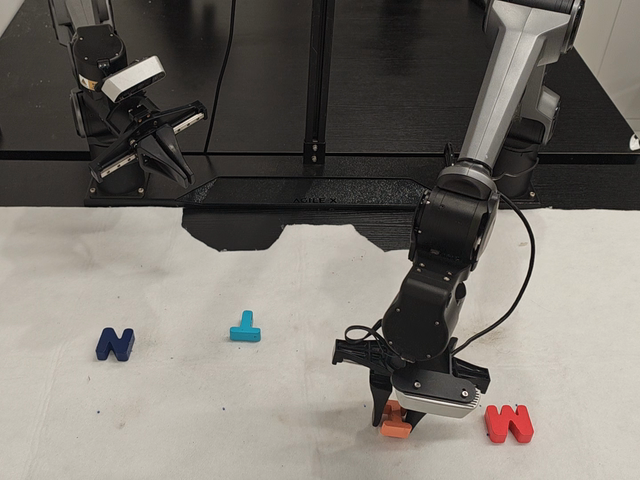}
& \includegraphics[width=.23\linewidth,trim=0 112 0 0,clip]{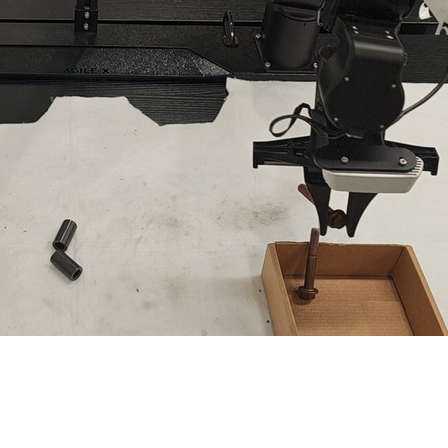}\\[3pt]
\includegraphics[width=.23\linewidth]{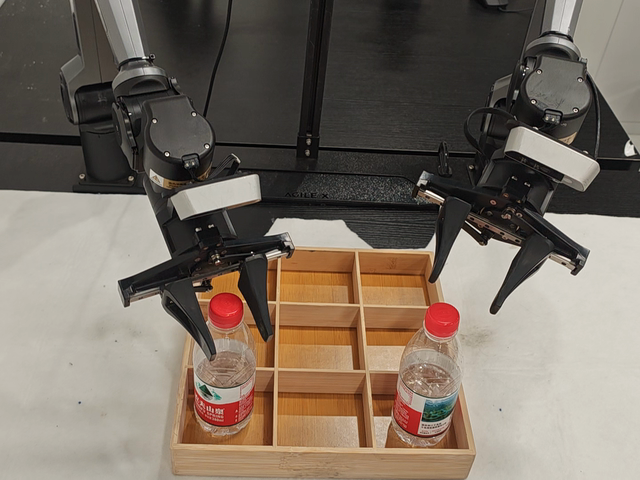}
& \includegraphics[width=.23\linewidth]{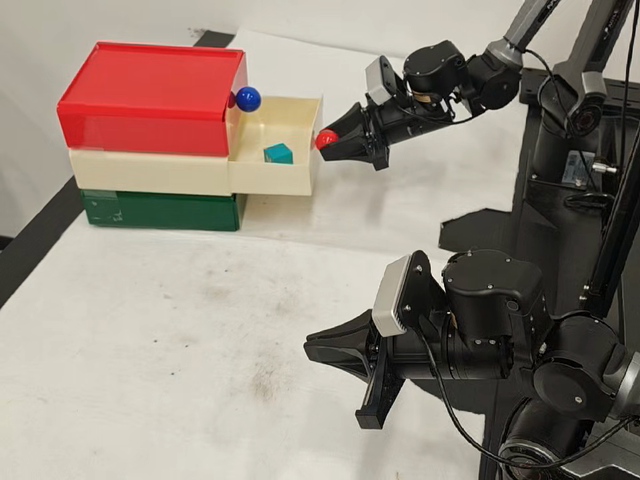}
\end{tabular}%
}
\raisebox{\depth}{\resizebox*{!}{\dimexpr\ht\realWorldTableBox+\dp\realWorldTableBox\relax}{\usebox{\realWorldImagesBox}}}%
\hspace{.035\linewidth}%
\raisebox{\depth}{\usebox{\realWorldTableBox}}
\caption{\textbf{Real-world manipulation results.}
Left: task examples, read row-wise in the order listed in the table.
Right: success rates of the $\pi_{0.5}$ baseline and $\pi_{0.5}$ + MorphAct.
Gains are absolute differences in percentage points (pp).
Each method is evaluated over 50 physical rollouts per task.}
\label{fig:real-world}
\end{figure}
\subsection{Real-World Manipulation}
\label{sec:real-world}

To evaluate MorphAct beyond simulation, we compare $\pi_{0.5}$ with
and without MorphAct on four physical tasks: Sort Letters,
Sort Hardware, Deliver Water, and Cabinet Storage.
These tasks involve arranging letter blocks, sorting hardware into
containers, transferring water bottles, and placing an object in a
cabinet, respectively. Each method is evaluated over 50 physical
rollouts per task.
Success requires completion of the specified manipulation objective.
Figure~\ref{fig:real-world} pairs representative rollout frames
with the success rates for each task.
MorphAct improves success on all four tasks by 8.00--14.00 percentage
points. The largest gain is on Sort Hardware, where success
increases from 8.00\% to 22.00\%; Sort Letters improves from
82.00\% to 92.00\%, Deliver Water from 78.00\% to 86.00\%,
and Cabinet Storage from 84.00\% to 96.00\%.
These results complement the simulation comparisons with improvements
during closed-loop execution on physical manipulation tasks.

\subsection{Inference Cost}
\label{sec:inference-cost}
MorphAct generates updates once per replanning context and reuses
them throughout action-chunk prediction, keeping parameter generation
outside the solver loop.
Under matched hardware and execution settings, end-to-end chunk
latency increases by approximately 18.00--21.00\% relative to shared LoRA
on $\pi_0$ and $\pi_{0.5}$.
This measurement covers observation preprocessing and complete
action-chunk prediction, including parameter generation and
low-rank residual computation.
Appendix~\ref{sec:adaptation-cost} reports generator parameter counts,
active-factor counts, and measured chunk latency, inference memory,
and control rate.

%% file: sections/analysis.tex
% Section source. See project README for the local-to-Overleaf editing workflow.
\section{Analysis}
\label{sec:analysis}

We examine three design choices in MorphAct: the condition used for
parameter generation, the modules receiving updates, and the features
connecting the two generators. Ablations use X-VLA on LIBERO under
the task setup and evaluation protocol in
Section~\ref{sec:eval-setting}.
Table~\ref{tab:ablations} reports success rates, with superscripts
linking each variant to the comparisons below.

\begin{table}[tbp]
\centering
\normalsize
\setlength{\tabcolsep}{3pt}
\renewcommand{\arraystretch}{1.15}
\begin{tabular}{
    @{}>{\raggedright\arraybackslash}p{.28\linewidth}
    *{5}{S[
        table-format=2.2,
        detect-weight=true,
        table-column-width=\dimexpr.144\linewidth-2\tabcolsep\relax
    ]}@{}
}
\toprule
Variant
& \multicolumn{1}{c}{Spatial $\uparrow$}
& \multicolumn{1}{c}{Object $\uparrow$}
& \multicolumn{1}{c}{Goal $\uparrow$}
& \multicolumn{1}{c}{Long $\uparrow$}
& \multicolumn{1}{c}{Mean $\uparrow$} \\
\midrule
\rowcolor{MorphActRow}
MorphAct\textsuperscript{A,C}
& \textbf{98.67} & \textbf{99.87} & \textbf{97.07}
& \textbf{93.20} & \textbf{97.20} \\
\midrule
Constant\textsuperscript{A}
& 93.40 & 90.00 & 89.00 & 77.60 & 87.50 \\
Task-ID\textsuperscript{A}
& 90.00 & 90.80 & 91.20 & 81.00 & 88.25 \\
\addlinespace[3pt]
VLM-only\textsuperscript{B}
& 40.00 & 59.00 & 42.00 & 33.00 & 43.50 \\
Action-only\textsuperscript{B}
& 88.80 & 92.00 & 73.80 & 69.40 & 81.00 \\
\addlinespace[3pt]
Unadapted\textsuperscript{B,C}
& 91.00 & 97.80 & 83.60 & 71.60 & 86.00 \\
\bottomrule
\end{tabular}
\caption{
\textbf{X-VLA ablation success rates on LIBERO (\%).}
Superscripts identify generation conditions
(A, Section~\ref{sec:context-specialization}), adapted modules
(B, Section~\ref{sec:module-coverage}), and feature coupling
(C, Section~\ref{sec:coordinated-adaptation}).
Mean averages the four suite scores.
Shading denotes MorphAct; bold marks column maxima.
}
\label{tab:ablations}
\end{table}

\subsection{Context Beyond Task Identity}
\label{sec:context-specialization}

Comparison A varies the generation condition while retaining the
policy's normal image, language, and robot-state inputs.
Constant uses one fixed condition across tasks, whereas Task-ID
uses one per task. Both retain MorphAct's generators.
Task-ID raises mean success from 87.50\% to 88.25\%, improving
Object, Goal, and Long but reducing Spatial by 3.40 percentage points.
MorphAct reaches 97.20\%, exceeding Constant and Task-ID by
9.70 and 8.95 points on average, with improvements on every suite.
The largest gains occur on Long, at 15.60 and 12.20 points,
respectively.
Both fixed-condition variants trail the X-VLA baseline's
93.05\% mean success in Table~\ref{tab:performance}, by 5.55
and 4.80 points.
MorphAct instead lets updates respond to the current observation
and robot state.
The comparison supports contextual generation beyond assigning
a fixed update to each task.

\subsection{Adapting Both Modules}
\label{sec:module-coverage}

Comparison B examines the contribution of each adapted module.
VLM-only and Action-only disable generated updates in the other
module but retain both modules in the policy.
Action-only achieves 81.00\% mean success versus 43.50\% for VLM-only,
making action-side generation the stronger single-module choice.
We then compare Action-only with Unadapted to separate upstream
adaptation from feature coupling.
Both use pre-adaptation features to condition the action generator.
Unadapted additionally adapts the representation supplied to the
action predictor, raising mean success from 81.00\% to 86.00\%
with improvements on all four suites.
Upstream adaptation thus benefits action prediction even when
its output does not condition action updates.

\subsection{Coupling through Adapted Features}
\label{sec:coordinated-adaptation}

Comparison C tests the features supplied to the action generator.
Unadapted and MorphAct both generate updates for both modules
and use adapted features for action prediction.
Unadapted conditions action-side generation on pre-adaptation
features, while MorphAct uses the adapted VLM output.
This raises mean success from 86.00\% to 97.20\%, a gain of
11.20 percentage points.
The suite-level results distinguish this effect from upstream
adaptation alone.
On Object, adding VLM updates raises success from Action-only's
92.00\% to Unadapted's 97.80\%, and coupling adds another 2.07 points.
On Long, upstream adaptation adds 2.20 points, while coupling
adds 21.60 points.
Coupling also improves Goal by 13.47 points.
Both feature sources support high success on Object, but Goal
and Long benefit more from generating action updates that respond
to the representation produced by upstream adaptation.

\textbf{Summary.}
The ablations support current-context conditioning, adaptation
of both modules, and coupling through adapted VLM features.
Task identity alone provides a small average gain over constant
conditioning, whereas current-context generation improves all four suites.
The progression from Action-only to Unadapted to MorphAct separates
the benefit of adapting the representation from the further benefit
of using it to generate action updates.
Together, these comparisons show that parameter generation
depends on both the current context and the adapted features used
for control.

%% file: sections/related_work.tex
% Section source. See project README for the local-to-Overleaf editing workflow.
\section{Related Work}

\textbf{VLA models.}
Generalist robot policies scale learning across tasks and embodiments
\citep{brohan2023rt1,openx2024}, while Octo supports adaptation to
new observation and action spaces \citep{octo2024}.
VIMA specifies manipulation tasks through interleaved textual and
visual prompts \citep{jiang2023vima}.
RT-2, OpenVLA, and the $\pi$ family connect pretrained vision--language
representations to robot actions
\citep{brohan2023rt2,kim2025openvla,black2025pi0,intelligence2025pi05}.
MorphAct studies how representation and action modules can adapt
to changing manipulation contexts within a shared policy.

\textbf{Parameter-efficient fine-tuning.}
Adapters and prefix-tuning learn small task-specific components
while freezing pretrained weights
\citep{houlsby2019adapters,li2021prefix}.
LoRA learns low-rank weight updates \citep{hu2022lora},
and DoRA separates weight magnitude and direction, applying
low-rank adaptation to the latter \citep{liu2024dora}.
For VLA models, OpenVLA-OFT studies efficient decoding and
fine-tuning objectives \citep{kim2025oft},
X-VLA uses embodiment-specific soft prompts \citep{zheng2026xvla},
and VLA-Adapter introduces Bridge Attention between vision--language
features and a lightweight action policy \citep{wang2026vlaadapter}.
MorphAct generates context-dependent low-rank updates while
retaining the host policy's action interface and objective.

\textbf{Parameter sharing and generation.}
Multi-task learning studies structural sharing
\citep{yang2020softmod,sun2022paco}, joint optimization
\citep{sener2018multitask,chen2018gradnorm,yu2020pcgrad,liu2021cagrad},
and task transfer and layerwise conflicts
\citep{fifty2021efficiently,shi2023recon}.
Parameter generation synthesizes network weights using
hypernetworks \citep{ha2017hypernetworks} or diffusion models
\citep{wang2024neural,wang2025scaling}, treating weights as
a generative modality \citep{wang2026position} and enabling
adaptation conditioned on tasks, features, examples, images,
or prompts \citep{mahabadi2021hyperformer,ivison2022hyperdecoders,
phang2023hypertuning,ruiz2024hyperdreambooth,liang2025drag}.
HY-WU studies structured synthesis of low-rank weight updates
\citep{wu2026hy}.
Hyper-LoRA generates instruction-conditioned updates
\citep{xiong2024efficient}, HyperVLA generates compact-policy parameters
once per episode for efficient inference \citep{xiong2025hypervla},
and PhaseLoRA modulates action-side LoRA per chunk
\citep{guo2026phaselora}.
MorphAct couples parameter generation across perception and action:
adapted vision--language representations condition action-side
parameter updates, and both generators are trained jointly
through the native action objective.

%% file: sections/conclusion.tex
\section{Conclusion}

MorphAct generates context-dependent low-rank updates for vision--language
and action modules through a shared adaptation mechanism.
Its generators are coupled through adapted representations and trained
jointly with the native action objective.
Atomic-task diagnostics reveal conflicting adaptation demands in both
modules, while evaluations on LIBERO, Meta-World, and four real-world
tasks demonstrate improved closed-loop manipulation.
Ablations further support contextual generation, adaptation of both
modules, and coupling through adapted VLM features.

%% file: sections/acknowledgments.tex
\section{Acknowledgments}

We thank Yong Zhong (\href{mailto:yzhong1024@gmail.com}{\texttt{yzhong1024@gmail.com}}) for technical support.

%% file: sections/appendix.tex
% Section source. See project README for the local-to-Overleaf editing workflow.
\clearpage
\appendix
\numberwithin{equation}{section}
\renewcommand{\theequation}{\thesection\arabic{equation}}
\renewcommand{\theHequation}{appendix.\thesection.\arabic{equation}}
\section{Atomic-Task Diagnostic Protocol and Complete Results}
\label{app:diagnostics}

The diagnostic uses the 65 native atomic tasks in RoboCasa365
\citep{nasiriany2026robocasa365}, as described in
Section~\ref{sec:diagnosis}. Each task identifier defines one group;
the complete identifiers appear in Table~\ref{tab:diag-source-ids}.
Task labels organize the diagnostic comparisons and are not inputs or
supervision for MorphAct.

\renewcommand{\thetable}{A\arabic{table}}
\renewcommand{\theHtable}{appendix.\arabic{table}}
\setcounter{table}{0}

\subsection{Parameter Scope and Sampling}
\label{app:diagnostic-sampling}

All three measurements start from the same RoboCasa-pretrained $\pi_{0.5}$ checkpoint and its associated normalization statistics. 
The VLM diagnostic set contains 2,923,335,408 scalar parameters, and the action set contains 430,098,464. These sets include vision and language backbones, token embeddings, the action expert, and their associated projections and normalization parameters. 
They are the full diagnostic sets, not merely the projections targeted by the generated low-rank updates. Token-embedding gradients retain their sparse representation.

For each diagnostic subset and sampling round, the sampler constructs $U_i$ and $V_i$ from disjoint episode groups. For alignment (A) and contribution accounting (C), each batch contains 32 samples from eight episodes, with four sampled timesteps per episode. Cross-task comparisons use $U_i,U_j$ with $i<j$; within-task controls use $U_i,V_i$. The 16 rounds yield $16\binom{65}{2}=33{,}280$ cross-task pair-rounds and $16\times65=1{,}040$ within-task observations per module. These are repeated batch samples at one checkpoint, not independent training seeds.

\begin{table}[htbp]
\centering\normalsize
\setlength{\tabcolsep}{4pt}
\renewcommand{\arraystretch}{1.13}
\begin{tabularx}{\linewidth}{@{}>{\raggedright\arraybackslash}p{0.235\linewidth}>{\raggedright\arraybackslash}X>{\raggedright\arraybackslash}X>{\raggedright\arraybackslash}X@{}}
\toprule
Setting & A: alignment & B: one-step affinity & C: contributions\\
\midrule
Batch size & 32 & 16 & 32\\
Compared batches & $U_i,U_j$; $U_i,V_i$ & Source $U_i$, target $V_j$ & $U_i,U_j$; $U_i,V_i$\\
Parameter precision & BF16 & FP32 from BF16 checkpoint & BF16\\
Sampling rounds & 16 & 4 & 16\\
Cross-task observations & 33,280 unordered pair-rounds & 16,640 directed pair-rounds & 33,280 unordered pair-rounds\\
Within-task observations & 1,040 & 260 & 1,040\\
Measurement & Module gradient cosine & Full-parameter update, target loss & 66 parameter groups\\
\bottomrule
\end{tabularx}
\caption{\textbf{Diagnostic settings.} A/C resample batches over 16 rounds; B uses four complete source-to-target rounds at the same checkpoint.}
\label{tab:diag-settings}
\end{table}

A uses microbatches of four, FP32 accumulation, and BF16 gradient storage. C independently computes group statistics on the same sample manifests, accumulating eight microbatch contributions per batch. Its full-module quantities are reconstructed from its own dot products and norms, so A and C have separate numerical realizations. Negative-cosine frequencies count observations within individual rounds; the sign of a pair's 16-round mean is a different statistic. Confidence intervals and zero-norm exclusion counts are not reported.

\subsection{Gradient Alignment and Directional Effects}

For module $m\in\{v,a\}$, we compute gradients of the native action
loss on each source batch and compare their alignment:
\begin{equation}
    g_i^{(m)}=\nabla_{\theta_m}\mathcal L(U_i;\theta),
    \qquad
    c_{ij}^{(m)}=
    \frac{\langle g_i^{(m)},g_j^{(m)}\rangle}
    {\lVert g_i^{(m)}\rVert_2\lVert g_j^{(m)}\rVert_2}.
    \label{eq:diagnostic-gradient}
\end{equation}
Negative alignment indicates competing local descent directions. The
conflict frequency is $p_m^-=\operatorname{mean}\mathbf{1}[c_{ij}^{(m)}<0]$.
Cross-task comparisons use $(U_i,U_j)$ with $i<j$; within-task
comparisons use independently sampled $(U_i,V_i)$.

Table~\ref{tab:diag-alignment} reports both conflict frequency $p_m^-$ and angular magnitude $b_m=\operatorname{mean}\max(0,-c^{(m)})$. The last column counts pairs with negative 16-round mean cosine, whereas $p_m^-$ counts negative observations within rounds.

\begin{table}[htbp]
\centering\normalsize
\setlength{\tabcolsep}{4pt}
\renewcommand{\arraystretch}{1.12}
\begin{tabular}{@{}llrrrr@{}}
\toprule
Module & Comparison & \shortstack{Mean cosine\\($10^{-3}$)} & $p_m^-$ (\%) & \shortstack{$b_m$\\($10^{-3}$)} & Negative mean pairs\\
\midrule
VLM & Cross-task & 1.05 & 47.80 & 7.00 & 890/2,080\\
VLM & Within-task & 10.24 & 38.94 & 12.01 & 11/65\\
Action & Cross-task & 3.00 & 48.28 & 26.76 & 918/2,080\\
Action & Within-task & 25.11 & 42.69 & 34.50 & 19/65\\
\bottomrule
\end{tabular}
\caption{\textbf{Gradient alignment at the fixed checkpoint.} Within-task batches come from disjoint episode groups. Mean negative part includes zeros from nonnegative observations.}
\label{tab:diag-alignment}
\end{table}

For one-step transfer, each source update starts from the common
checkpoint. Following the affinity convention of \citet{fifty2021efficiently},
we define
\begin{equation}
    \theta_i^+=\theta-\eta\nabla_\theta\mathcal L(U_i;\theta),
    \qquad
    Z_{i\rightarrow j}
    =1-\frac{\mathcal L(V_j;\theta_i^+)}
             {\mathcal L(V_j;\theta)}.
    \label{eq:diagnostic-transfer}
\end{equation}
Positive $Z_{i\rightarrow j}$ means the update reduces target loss;
negative values mean it increases target loss.

For a sufficiently small unpreconditioned update, $\mathcal L(V_j;\theta-\eta g_i)-\mathcal L(V_j;\theta)=-\eta\langle\nabla_\theta\mathcal L(V_j;\theta),g_i\rangle+O(\eta^2)$, where $g_i=\nabla_\theta\mathcal L(U_i;\theta)$. This first-order relation motivates the gradient diagnostic; the directional experiment measures the finite-step loss change directly.

B selects 16 samples per side at positions $4k$ and $4k+1$, $k=0,\ldots,7$, from each original 32-sample batch. Source gradients accumulate in FP32 with microbatches of four; target evaluation uses a microbatch of 16. Each source update starts afresh from the common checkpoint and applies Equation~\eqref{eq:diagnostic-transfer} with $\eta\approx0.011322$, calibrated using the first three source identifiers. Parameters are converted from the BF16 checkpoint to FP32 for this experiment. Target observations, action targets, flow times, and noise are held fixed before and after each update. Source self-improvement is measured on $U_i$, not on the independently sampled diagonal target $V_i$.

\begin{table}[htbp]
\centering\normalsize
\setlength{\tabcolsep}{5pt}
\renewcommand{\arraystretch}{1.12}
\begin{tabular}{@{}lrrrrr@{}}
\toprule
Comparison & Loss up & Loss down & Unchanged & \shortstack{Mean $Z$\\($10^{-3}\%$)} & \shortstack{Mean increase\\($10^{-3}\%$)}\\
\midrule
Cross-task & 8,310 & 8,326 & 4 & $-0.47$ & 44.36\\
Within-task & 133 & 127 & 0 & $+4.89$ & 96.62\\
\bottomrule
\end{tabular}
\caption{\textbf{One-step target effects across four sampling rounds.} Affinity and loss changes are relative to the target's initial loss. Mean increase is conditional on a loss-increase event. A positive $Z$ means lower target loss.}
\label{tab:diag-affinity}
\end{table}

The stored relative-loss-increase matrix $H$ is converted to affinity only by $Z=-H$ (\citealp{fifty2021efficiently}). Table~\ref{tab:diag-affinity} pools all four complete $65\times65$ evaluations. All 260 source updates reduce source-batch loss, by 2.38\% on average (range 0.37--16.61\%). Each source update harms 22--46 other target batches, with a median of 32/64; conversely, each target receives antagonistic updates from at least 20 other sources within its round. The maximum cross-task relative loss increase is 0.80\%. These are local loss effects, not closed-loop success rates.

\subsection{Parameter Groups and Contribution Accounting}
\label{app:diagnostic-groups}

The parameter partition has 47 VLM groups: 27 vision blocks, 18 language blocks, token embeddings, and one group outside Transformer blocks. The action set has 19 groups: 18 Transformer blocks and one outside-block group. Only actual scanned Transformer axes are split by depth. The action outside-block group includes final conditional-normalization mapping, action input/output projections, and the time MLP. The VLM counterpart includes image input/output mappings, position embeddings, and final normalization parameters. These groups remain distinct from Transformer layers.

For a fixed module and sampling round, group $\ell$
contributes to the full-module alignment as
\begin{equation}
\begin{aligned}
    q_{ij}^{(\ell)}
    &=\frac{\langle g_i^{(\ell)},g_j^{(\ell)}\rangle}
            {\lVert g_i\rVert_2\lVert g_j\rVert_2},
    & c_{ij}&=\sum_\ell q_{ij}^{(\ell)}=P_{ij}-N_{ij},\\
    P_{ij}&=\sum_\ell\max(q_{ij}^{(\ell)},0),
    & N_{ij}&=\sum_\ell\max(-q_{ij}^{(\ell)},0).
\end{aligned}
\label{eq:diagnostic-contributions}
\end{equation}
The full-module denominator makes the group contributions additive.
We quantify their cancellation by pooling absolute contribution mass
over pair-rounds:
\begin{equation}
    C_m
    =1-
    \frac{\sum_{r,i<j}
        \left|P_{rij}^{(m)}-N_{rij}^{(m)}\right|}
         {\sum_{r,i<j}
        \left(P_{rij}^{(m)}+N_{rij}^{(m)}\right)}.
    \label{eq:diagnostic-cancellation}
\end{equation}

For each pair and round, Equation~\eqref{eq:diagnostic-contributions} gives $c=P-N$. The cancelled mass is $P+N-|P-N|$ and is computed \emph{before} averaging across pairs or rounds. Equation~\eqref{eq:diagnostic-cancellation} divides its pooled sum by total absolute contribution mass. Cancellation therefore measures opposing normalized contributions, not the proportion of conflicting parameters or a percentage loss degradation.

\begin{table}[htbp]
\centering\normalsize
\setlength{\tabcolsep}{6pt}
\renewcommand{\arraystretch}{1.12}
\begin{tabular}{@{}llrrr@{}}
\toprule
Module & Comparison & Observations & Mixed signs (\%) & Cancellation (\%)\\
\midrule
VLM & Cross-task & 33,280 & 99.45 & 36.13\\
VLM & Within-task & 1,040 & 94.13 & 23.51\\
Action & Cross-task & 33,280 & 95.93 & 30.90\\
Action & Within-task & 1,040 & 90.10 & 22.70\\
\bottomrule
\end{tabular}
\caption{\textbf{Opposing parameter-group contributions.} Mixed-sign observations satisfy both $P>0$ and $N>0$. Cancellation pools absolute contribution mass within observations.}
\label{tab:diag-cancellation}
\end{table}

Groups are ranked by their share of negative cross-task contribution, $\sum_{r,i<j}\max(-q_{rij}^{(\ell)},0)$ divided by the sum over groups in the same module. The three largest action groups---outside Transformer blocks (39.02\%), block 00 (23.74\%), and block 11 (16.11\%)---account for 78.88\% before rounding. The largest VLM groups are outside Transformer blocks (20.40\%), language block 00 (17.44\%), and vision block 00 (7.83\%). Table~\ref{tab:diag-rankings} lists all groups. Depth profiles use group-local cosine with group-specific gradient norms; the additive contribution $q$ instead uses full-module norms.

\begin{table}[!htbp]
\centering\normalsize
\setlength{\tabcolsep}{8pt}
\renewcommand{\arraystretch}{1.08}
\begin{tabular}{@{}rlrrr@{}}
\toprule
Rank & VLM parameter group & Params (M) & Conflict (\%) & Negative share (\%)\\
\midrule
1 & Outside Transformer blocks & 3.34 & 48.56 & 20.40\\
2 & Language block 00 & 110.11 & 48.93 & 17.44\\
3 & Vision block 00 & 15.24 & 48.15 & 7.83\\
4 & Vision block 01 & 15.24 & 48.23 & 3.91\\
5 & Language block 14 & 110.11 & 48.86 & 3.48\\
6 & Language block 07 & 110.11 & 48.80 & 3.14\\
7 & Vision block 02 & 15.24 & 48.08 & 3.00\\
8 & Language block 09 & 110.11 & 48.73 & 2.98\\
9 & Language block 06 & 110.11 & 49.26 & 2.95\\
10 & Language block 01 & 110.11 & 48.50 & 2.76\\
11 & Language block 04 & 110.11 & 49.65 & 2.58\\
12 & Language block 03 & 110.11 & 49.25 & 2.28\\
13 & Vision block 04 & 15.24 & 48.07 & 2.28\\
14 & Vision block 03 & 15.24 & 47.94 & 2.00\\
15 & Language block 05 & 110.11 & 49.50 & 1.96\\
16 & Language block 11 & 110.11 & 48.79 & 1.89\\
17 & Language block 08 & 110.11 & 48.81 & 1.67\\
18 & Language block 10 & 110.11 & 48.65 & 1.63\\
19 & Language block 02 & 110.11 & 49.63 & 1.36\\
20 & Language block 13 & 110.11 & 48.59 & 1.35\\
21 & Language block 12 & 110.11 & 48.80 & 1.28\\
22 & Vision block 06 & 15.24 & 47.67 & 1.07\\
23 & Token embedding & 526.65 & 48.97 & 1.00\\
24 & Vision block 05 & 15.24 & 47.60 & 0.91\\
25 & Vision block 26 & 15.24 & 48.50 & 0.70\\
26 & Vision block 08 & 15.24 & 47.62 & 0.68\\
27 & Vision block 24 & 15.24 & 48.75 & 0.64\\
28 & Vision block 07 & 15.24 & 47.82 & 0.62\\
29 & Vision block 09 & 15.24 & 48.11 & 0.46\\
30 & Vision block 12 & 15.24 & 47.94 & 0.46\\
31 & Vision block 13 & 15.24 & 48.30 & 0.45\\
32 & Vision block 14 & 15.24 & 48.40 & 0.43\\
33 & Vision block 10 & 15.24 & 48.00 & 0.42\\
\bottomrule
\end{tabular}
\caption{\textbf{Complete parameter-group rankings.} Groups are ranked within each module by negative cross-task contribution share. Conflict frequencies use group-local cosines. The remaining 14 VLM groups and all 19 action groups continue on the next page.}
\label{tab:diag-rankings}
\end{table}

\begin{table}[!htbp]
\centering\normalsize
\textbf{Table~\ref{tab:diag-rankings} (continued): complete parameter-group rankings.}\\[5pt]
\setlength{\tabcolsep}{8pt}
\renewcommand{\arraystretch}{1.08}
\begin{tabular}{@{}rlrrr@{}}
\toprule
Rank & Parameter group & Params (M) & Conflict (\%) & Negative share (\%)\\
\midrule
\multicolumn{5}{@{}l}{\textit{Vision--language module (continued)}}\\
34 & Vision block 11 & 15.24 & 48.04 & 0.40\\
35 & Vision block 25 & 15.24 & 48.64 & 0.38\\
36 & Vision block 23 & 15.24 & 49.00 & 0.37\\
37 & Language block 17 & 110.11 & 49.52 & 0.35\\
38 & Vision block 15 & 15.24 & 48.47 & 0.31\\
39 & Language block 16 & 110.11 & 49.03 & 0.27\\
40 & Vision block 17 & 15.24 & 48.55 & 0.27\\
41 & Vision block 16 & 15.24 & 48.34 & 0.25\\
42 & Vision block 22 & 15.24 & 49.18 & 0.25\\
43 & Vision block 18 & 15.24 & 48.56 & 0.25\\
44 & Vision block 21 & 15.24 & 48.58 & 0.24\\
45 & Vision block 20 & 15.24 & 49.13 & 0.24\\
46 & Vision block 19 & 15.24 & 48.31 & 0.22\\
47 & Language block 15 & 110.11 & 49.64 & 0.19\\
\midrule
\multicolumn{5}{@{}l}{\textit{Action module}}\\
1 & Outside Transformer blocks & 5.32 & 48.22 & 39.02\\
2 & Action block 00 & 23.60 & 49.29 & 23.74\\
3 & Action block 11 & 23.60 & 49.53 & 16.11\\
4 & Action block 17 & 23.60 & 48.53 & 4.48\\
5 & Action block 01 & 23.60 & 48.84 & 4.42\\
6 & Action block 16 & 23.60 & 48.45 & 2.30\\
7 & Action block 02 & 23.60 & 48.46 & 1.97\\
8 & Action block 14 & 23.60 & 48.44 & 1.50\\
9 & Action block 08 & 23.60 & 49.12 & 0.93\\
10 & Action block 07 & 23.60 & 48.16 & 0.78\\
11 & Action block 15 & 23.60 & 48.87 & 0.71\\
12 & Action block 09 & 23.60 & 48.19 & 0.70\\
13 & Action block 04 & 23.60 & 48.59 & 0.64\\
14 & Action block 12 & 23.60 & 48.22 & 0.64\\
15 & Action block 13 & 23.60 & 48.28 & 0.58\\
16 & Action block 10 & 23.60 & 47.94 & 0.53\\
17 & Action block 03 & 23.60 & 48.13 & 0.47\\
18 & Action block 06 & 23.60 & 48.51 & 0.33\\
19 & Action block 05 & 23.60 & 47.76 & 0.15\\
\bottomrule
\end{tabular}
\end{table}

\begin{table}[!htbp]
\centering\normalsize
\setlength{\tabcolsep}{5pt}
\renewcommand{\arraystretch}{1.1}
\begin{tabular}{@{}rlrl@{}}
\toprule
ID & Original RoboCasa source-task identifier & ID & Original RoboCasa source-task identifier\\
\midrule
1 & AdjustToasterOvenTemperature & 34 & PickPlaceCounterToCabinet\\
2 & AdjustWaterTemperature & 35 & PickPlaceCounterToDrawer\\
3 & CheesyBread & 36 & PickPlaceCounterToMicrowave\\
4 & CloseBlenderLid & 37 & PickPlaceCounterToOven\\
5 & CloseCabinet & 38 & PickPlaceCounterToSink\\
6 & CloseDishwasher & 39 & PickPlaceCounterToStandMixer\\
7 & CloseDrawer & 40 & PickPlaceCounterToStove\\
8 & CloseElectricKettleLid & 41 & PickPlaceCounterToToasterOven\\
9 & CloseFridge & 42 & PickPlaceDrawerToCounter\\
10 & CloseFridgeDrawer & 43 & PickPlaceFridgeDrawerToShelf\\
11 & CloseMicrowave & 44 & PickPlaceFridgeShelfToDrawer\\
12 & CloseOven & 45 & PickPlaceMicrowaveToCounter\\
13 & CloseStandMixerHead & 46 & PickPlaceSinkToCounter\\
14 & CloseToasterOvenDoor & 47 & PickPlaceStoveToCounter\\
15 & CoffeeServeMug & 48 & PickPlaceToasterOvenToCounter\\
16 & CoffeeSetupMug & 49 & PickPlaceToasterToCounter\\
17 & LowerHeat & 50 & PreheatOven\\
18 & MakeIcedCoffee & 51 & SlideDishwasherRack\\
19 & NavigateKitchen & 52 & SlideOvenRack\\
20 & OpenBlenderLid & 53 & SlideToasterOvenRack\\
21 & OpenCabinet & 54 & StartCoffeeMachine\\
22 & OpenDishwasher & 55 & TurnOffMicrowave\\
23 & OpenDrawer & 56 & TurnOffSinkFaucet\\
24 & OpenElectricKettleLid & 57 & TurnOffStove\\
25 & OpenFridge & 58 & TurnOnBlender\\
26 & OpenFridgeDrawer & 59 & TurnOnElectricKettle\\
27 & OpenMicrowave & 60 & TurnOnMicrowave\\
28 & OpenOven & 61 & TurnOnSinkFaucet\\
29 & OpenStandMixerHead & 62 & TurnOnStove\\
30 & OpenToasterOvenDoor & 63 & TurnOnToaster\\
31 & PackDessert & 64 & TurnOnToasterOven\\
32 & PickPlaceCabinetToCounter & 65 & TurnSinkSpout\\
33 & PickPlaceCounterToBlender & &\\
\bottomrule
\end{tabular}
\caption{\textbf{Original identifiers retained in the diagnostic manifests.} IDs follow the original alphabetical order. Each ID denotes one native RoboCasa365 atomic task and one diagnostic group.}
\label{tab:diag-source-ids}
\end{table}

\clearpage
\subsection{Parameter-Group Visualizations}

\begin{figure}[htbp]
\centering
\includegraphics[width=\linewidth]{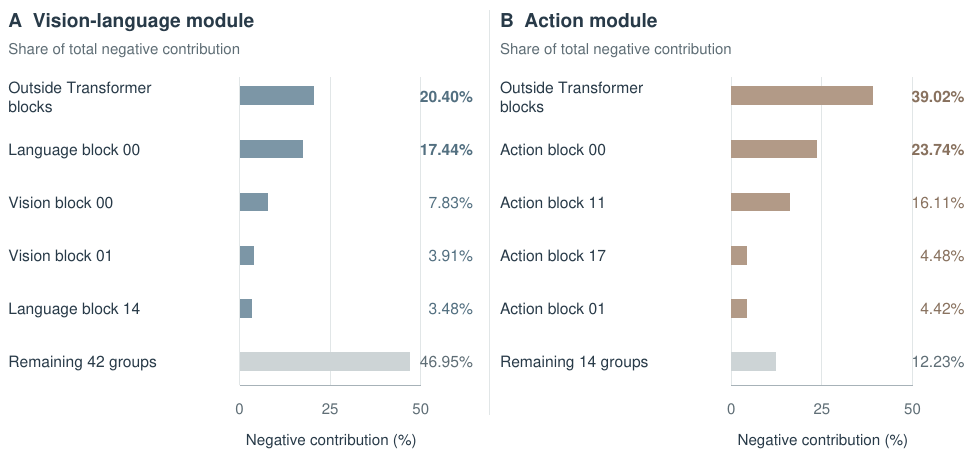}
\caption{\textbf{Negative contributions are concentrated.} The five largest groups are shown individually for each module; all others are explicitly combined. Parameters outside Transformer blocks remain separately identified. Percentages are shares of pooled negative cross-task contribution, not fractions of model parameters.}
\label{fig:diagnostic-contributions}
\end{figure}

\begin{figure}[!htbp]
\centering
\includegraphics[width=\linewidth]{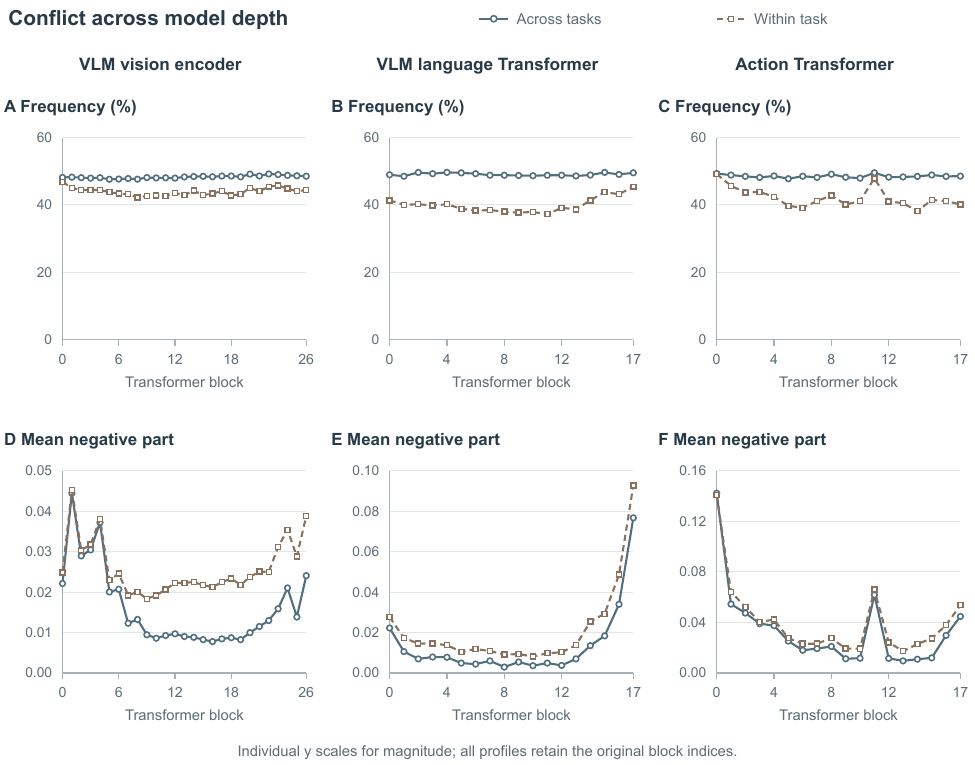}
\caption{\textbf{Complete Transformer depth profiles.} Cross- and within-task comparisons report negative-cosine frequency and the mean negative part of group-local cosine. Block indices follow the actual 27-block vision, 18-block language, and 18-block action layouts.}
\label{fig:diagnostic-depth}
\end{figure}

Figure~\ref{fig:diagnostic-contributions} retains the five largest negative-contribution groups in each module and explicitly aggregates the rest. Figure~\ref{fig:diagnostic-depth} reports complete depth profiles for the vision, language, and action Transformers. Frequencies and mean negative parts use group-local cosine; contribution shares use the full-module-normalized decomposition. These distinct normalizations prevent a large group-local angular conflict from being mistaken for a large contribution to the module's total interference.

\clearpage

% Section source. See project README for the local-to-Overleaf editing workflow.
\section{Parameter Geometry and Architecture Interfaces}
\label{app:parameter-geometry}

With $H_q$ query heads of width $d_h$, the update is assembled from output-head blocks:

\begin{equation}
\begin{aligned}
\Delta W_Q&=\left[
\frac{\alpha}{r}A_1B_1\;\middle|\;\cdots\;\middle|\;
\frac{\alpha}{r}A_{H_q}B_{H_q}\right],\\
A_h&\in\mathbb R^{d_{\mathrm{in}}\times r},
\qquad B_h\in\mathbb R^{r\times d_h}.
\end{aligned}
\end{equation}

Key and value projections use their actual number of key--value heads. For the attention output projection, input-head contributions are summed:

\begin{equation}
\begin{aligned}
\delta y_O&=\frac{\alpha}{r}\sum_{h=1}^{H_q}u_hA_hB_h,\\
A_h&\in\mathbb R^{d_h\times r},
\qquad B_h\in\mathbb R^{r\times d_{\mathrm{out}}}.
\end{aligned}
\end{equation}

Each factor product has rank at most $r$; an assembled projection can reach the sum of block ranks, bounded by its dimensions. Matrix-level and output-head constructions have scalar factor counts

\begin{equation}
N_{\mathrm{matrix}}=r(d_{\mathrm{in}}+d_{\mathrm{out}}),
\end{equation}

and

\begin{equation}
N_{\mathrm{heads}}=H_qr(d_{\mathrm{in}}+d_h),
\end{equation}

respectively. These are per-projection, per-context output budgets, excluding generator weights. Equal block rank therefore does not imply equal generated or trainable capacity. The code validates layer and head counts, projection widths, factor shapes, and scales before residual injection.

The general interface in Equation~\eqref{eq:morphact-generation} accommodates different base policies. \tablename~\ref{tab:interfaces} summarizes the integrations. X-VLA retains its fused QKV base projections while adding separate generated Q/K/V residuals. Its action generator additionally uses auxiliary-view visual features; this optional extension does not apply to the $\pi$ instantiations.

\begin{table}[htbp]
\centering
\normalsize
\setlength{\tabcolsep}{3pt}
\renewcommand{\arraystretch}{1.15}
\begin{tabularx}{\linewidth}{@{}>{\raggedright\arraybackslash}X>{\raggedright\arraybackslash}X>{\raggedright\arraybackslash}X>{\raggedright\arraybackslash}X@{}}
\toprule
Backbone & VLM-generator condition / targets & Action-generator condition / targets & Training prediction \\
\midrule
$\pi_0$ / $\pi_{0.5}$ & Multimodal prefix embeddings / selected VLM projections & Adapted final tokens and state / selected action-expert projections & Flow velocity \\
X-VLA & Visual/language embeddings / selected Florence encoder projections & Adapted features, auxiliary views, state / selected action-Transformer projections & Clean action \\
\bottomrule
\end{tabularx}
\caption{\textbf{Architecture-specific interfaces.} Both integrations preserve the base policy's action-conditioning and loss definitions. Additional view and domain interfaces are specific to X-VLA.}
\label{tab:interfaces}
\end{table}

% Section source. See project README for the local-to-Overleaf editing workflow.
\Needspace{10\baselineskip}

% Replace the opening of the existing appendix section and its two
% training/inference procedures with this block.
% Keep the subsequent objective, initialization, precision, and
% variable-length conditioning paragraphs.
%
% Required in the main-file preamble:
% \usepackage{algorithm}
% \usepackage{algpseudocode}

\section{Joint Training and Policy Execution}
\label{app:training-inference}

Algorithms~\ref{alg:morphact-training} and
\ref{alg:morphact-inference} summarize one training update and
action-chunk prediction at one replanning context.
Both follow the generation order in
Equation~\eqref{eq:morphact-generation}: generate the VLM update,
compute the adapted representation, and generate the action update.
Generated updates are applied as low-rank residuals, without
materializing dense weight updates.

Only $\Phi=(\phi_v,\phi_a,\rho_a,\psi)$ is optimized, where
$\psi$ denotes the projector parameters retained as trainable
by the adaptation configuration. The remaining backbone
parameters stay frozen. The condition constructor $C_a$ uses
adapted features and robot state; for X-VLA, it additionally
includes native auxiliary-view features.
Training evaluates per-context operations over the demonstration batch.

\begin{algorithm}[H]
\caption{MorphAct: One Joint Training Update}
\label{alg:morphact-training}
\normalsize
\algrenewcommand{\alglinenumber}[1]{\normalsize #1:}
\begin{algorithmic}[1]
\Require Demonstration batch $\mathcal B=\{(x_t,Y_t)\}$,
         with $x_t=(o_t,\ell,s_t)$;
         parameters $\theta,\Phi$;
         optimizer
\Ensure Updated trainable parameters $\Phi$

\State Apply native preprocessing and construct embeddings
       $e_t=E(o_t,\ell)$ and validity masks.
\State $\Delta\theta_{v,t}\gets G_v(e_t;\phi_v)$
\State $(h_t,C_t)\gets
       F_v(e_t;\theta_v+\Delta\theta_{v,t})$
\State Construct $u_{a,t}$ using $C_a$ from $h_t$, $s_t$,
       and any native auxiliary-view features.
\State $\Delta\theta_{a,t}\gets G_a(u_{a,t};\phi_a)$
       \Comment{Keep $h_t$ and $C_t$ differentiable}

\State Sample the native objective's random variables $\xi_t$.
\State Construct the native training inputs from $(Y_t,\xi_t)$.
\State Compute $\widehat y_t$ using the training inputs,
       $C_t$, $s_t$, and the action module with
       parameters $\theta_a+\Delta\theta_{a,t}$.
\State $\mathcal L\gets
       \operatorname{mean}_{(x_t,Y_t)\in\mathcal B}
       \mathcal L_{\mathrm{native}}(\widehat y_t,Y_t;\xi_t)$

\State $g_\Phi\gets\nabla_\Phi\mathcal L$
       \Comment{Through action conditioning and $h_t\to G_a$}
\If{$\mathcal L$ and $\lVert g_\Phi\rVert_2$ are finite}
    \State Apply global gradient clipping to $g_\Phi$.
    \State Update $\Phi$ using the configured optimizer and schedule.
\Else
    \State Skip the parameter update.
\EndIf
\State \Return $\Phi$
\end{algorithmic}
\end{algorithm}

The factors produced by $G_v$ and $G_a$ are differentiable
intermediate outputs, not separate optimization variables.
The native action loss trains the VLM generator through both
$C_t$ and the dependence of $\Delta\theta_{a,t}$ on $h_t$.
No operation labels or operation-specific adapter targets are used.
The prediction target follows the backbone: flow velocity for
$\pi_0$ and $\pi_{0.5}$, and clean actions for X-VLA.

\begin{algorithm}[H]
\caption{MorphAct: Action-Chunk Prediction at One Context}
\label{alg:morphact-inference}
\normalsize
\algrenewcommand{\alglinenumber}[1]{\normalsize #1:}
\begin{algorithmic}[1]
\Require Current context $x_t=(o_t,\ell,s_t)$;
         learned parameters $\theta,\Phi$;
         native action sampler
\Ensure Predicted action chunk $\widehat Y_t$

\State Apply native preprocessing and construct embeddings
       $e_t=E(o_t,\ell)$ and validity masks.
\State $\Delta\theta_{v,t}\gets G_v(e_t;\phi_v)$
\State $(h_t,C_t)\gets
       F_v(e_t;\theta_v+\Delta\theta_{v,t})$
\State Construct $u_{a,t}$ using $C_a$ from $h_t$, $s_t$,
       and any native auxiliary-view features.
\State $\Delta\theta_{a,t}\gets G_a(u_{a,t};\phi_a)$
\State Cache the generated updates and adapted condition $C_t$.

\State Initialize the backbone's native action sampler.
\For{each action-module evaluation required by the sampler}
    \State Compute the native prediction using the current sampler
           inputs, $C_t$, $s_t$, and
           $\theta_a+\Delta\theta_{a,t}$.
    \State Advance the sampler using its original update rule.
           \Comment{No parameter regeneration}
\EndFor
\State Obtain $\widehat Y_t$ from the completed sampling procedure.
\State \Return $\widehat Y_t$
\end{algorithmic}
\end{algorithm}

The surrounding control loop executes the first $H_{\mathrm{exec}}$
actions of $\widehat Y_t$ and acquires a new observation for replanning.
Both generated updates and the adapted conditioning states are
recomputed for that new context.
The generators do not receive noisy actions or internal solver time;
these remain inputs to the native action-prediction procedure.
The predicted horizon, executed-prefix length, and solver-step count
follow the evaluation configuration.

% Keep the existing appendix content from
% "Objective instantiations." onward below this point.

\textbf{Objective instantiations.} With $z_\tau=\tau\epsilon+(1-\tau)Y_t$, $\pi_0$ and $\pi_{0.5}$ minimize velocity-prediction error,

\begin{equation}
\mathcal L_\pi=\mathbb E_{\tau,\epsilon}
\left[\left\|v_\Phi(z_\tau,\tau,x_t)-(\epsilon-Y_t)\right\|_2^2\right],
\end{equation}

using the base implementation's reduction over action entries. The X-VLA integration instead predicts clean actions:

\begin{equation}
\mathcal L_{\mathrm{X\text{-}VLA}}=\mathbb E_{\tau,\epsilon}
\left[\ell_{\mathrm{action\text{-}space}}
\bigl(f_\Phi(z_\tau,\tau,x_t),Y_t\bigr)\right].
\end{equation}

Its action-space loss retains the original treatment of position, rotation, and gripper components. These instantiate Equation~\eqref{eq:morphact-objective}.

\textbf{Initialization and residual scale.} Both A/B readout weights are initialized from a zero-mean Gaussian with standard deviation $\sigma_0/\sqrt{d_g}$ and zero bias, where $d_g$ is the readout input width. The generated residual $\delta y=(\alpha/r)uAB$ obeys $\lVert\delta y\rVert_2\le(|\alpha|/r)\lVert u\rVert_2\lVert A\rVert_2\lVert B\rVert_2$, using spectral norms for the factors. Small factor scales thus favor small local residuals, without imposing an exactly zero initial policy update.

If an entire factor readout is initialized to zero, the generator trunk initially receives no gradient through that product: one factor derivative is multiplied by zero, while the zero readout has zero input Jacobian. Small, nonzero weights on both readouts avoid this structural blockage.

\textbf{Precision and optimization.} Generator parameters are stored in FP32; internal attention and feed-forward computation can use BF16. Factor readouts, low-rank multiplications, scaling, and residual accumulation are evaluated in FP32 before conversion to the base output dtype. The conversion can still quantize a sufficiently small residual. The training implementation uses one AdamW schedule with warmup and cosine decay, global gradient clipping, and checks that prevent updates on non-finite losses or gradient norms. The $\pi$ implementation additionally checks trainable-gradient coverage. Non-reentrant activation checkpointing reduces activation memory.

\textbf{Variable-length conditioning.} Valid condition tokens are compacted using the supplied masks, grouped by sequence length, and restored to the original batch order after generation. Token compaction and reordering preserve gradients to the condition features. Grouping can require separate generator invocations for different sequence lengths within a batch; generated updates are reused per observation and action chunk. Masks follow each integration's interface.

% Section source. See project README for the local-to-Overleaf editing workflow.
\Needspace{20\baselineskip}
\section{Additional Experimental Details}

\subsection{Ablation Configurations}

The variants in Table~\ref{tab:ablations} examine conditioning, adaptation
targets, and the representation passed between generators. Task-ID
assigns a fixed generation condition to each task, while Constant uses
conditions that remain fixed across tasks and observations. Both variants
retain the policy's normal image, language, and robot-state inputs.
VLM-only disables generated action-side updates; Action-only disables
generated VLM updates while retaining visual inputs. Unadapted keeps both
generators active but feeds the action generator the VLM representation
computed before the generated VLM update. The action predictor continues
to receive the adapted representation, so this variant specifically
changes the condition used for action-side parameter generation.

\Needspace{34\baselineskip}
\subsection{Adaptation Budget and Runtime}
\label{sec:adaptation-cost}

For $N_{\mathrm{solve}}$ action-refinement steps, the approximate computation is $T_{\mathrm{policy}}\approx T_E+T_{G_v}+T_{F_v}+T_{G_a}+N_{\mathrm{solve}}T_{F_a}$. Generator costs include condition construction, and adapted-module costs include the generated residuals. Generation is paid per context rather than per solver step.

Contextual parameterization separates learned generator weights from the updates instantiated for an observation. We report the parameter count of the two generators and, separately, the number of scalar A/B entries in an active low-rank update. Shared LoRA reuses learned factors across contexts, whereas MorphAct generates them from the current context; generated factors are outputs, not additional independently optimized parameters. Explicit factor counts also account for differences between head-aware and matrix-level layouts that a nominal rank alone does not capture.

Table~\ref{tab:cost} pairs these counts with end-to-end chunk latency, measured inference memory, and control rate. Timing covers observation preprocessing and complete action-chunk prediction, including parameter generation and residual computation where applicable. Methods are compared on matched hardware and execution settings, including solver steps, predicted action horizon, and executed-prefix length. Control rate is measured under the actual action-execution schedule, rather than inferred solely from the inverse of chunk-prediction latency.

\begin{table}[htbp]
\centering
\normalsize
\setlength{\tabcolsep}{3pt}
\renewcommand{\arraystretch}{1.15}
\begin{tabularx}{\linewidth}{
    @{}
    >{\raggedright\arraybackslash}p{0.23\linewidth}
    *{5}{>{\centering\arraybackslash}X}
    @{}
}
\toprule
Adaptation
& \shortstack{Generator\\params (M)}
& \shortstack{Active\\factors (M)}
& \shortstack{Chunk\\latency (ms)}
& \shortstack{Memory\\(GiB)}
& \shortstack{Rate\\(Hz)} \\
\midrule
\multicolumn{6}{@{}l}{\textit{Backbone: $\pi_0$}} \\
Shared LoRA
& -- & 49.99 & 120.00 & 7.43 & 28.09 \\
MorphAct
& 300.09 & 49.99 & 145.00 & 9.10 & 23.20 \\
\midrule
\multicolumn{6}{@{}l}{\textit{Backbone: $\pi_{0.5}$}} \\
Shared LoRA
& -- & 49.99 & 138.71 & 7.88 & 24.99 \\
MorphAct
& 300.09 & 49.99 & 164.00 & 9.53 & 21.20 \\
\bottomrule
\end{tabularx}
\caption{
\textbf{Adaptation capacity and execution cost.}
Generator counts include the two generators only;
``--'' indicates the absence of a generator.
Active factors count scalar A/B entries used by one context,
not the number of matrices.
Execution measurements use matched settings within each backbone.
Control rates are measured under the action-execution schedule.
}
\label{tab:cost}
\end{table}

MorphAct reuses the generated updates throughout action-chunk prediction, keeping parameter generation outside the solver loop. This amortizes generation across refinement steps but retains the cost of computing and applying the updates. Relative to shared LoRA, chunk latency rises by 25.00 ms on $\pi_0$ and 25.29 ms on $\pi_{0.5}$, while inference memory rises by 1.67 and 1.65 GiB, respectively. Measured control rates are 23.20 and 21.20 Hz for MorphAct, compared with 28.09 and 24.99 Hz for shared LoRA. Read together with Table~\ref{tab:performance}, these measurements characterize the manipulation-performance--execution-cost trade-off of contextual adaptation.